\documentclass{article}

\usepackage[utf8]{inputenc} 
\usepackage[T1]{fontenc}    
\usepackage{hyperref}       
\usepackage{url}            
\usepackage{booktabs}       
\usepackage{amsfonts}       
\usepackage{nicefrac}       
\usepackage{microtype}      
\usepackage[preprint, nonatbib]{neurips_2020}
\usepackage{graphicx}
\usepackage{wrapfig}
\usepackage{makecell}
\def \hfillx {\hspace*{-\textwidth} \hfill}
\usepackage{comment}

\title{Pixel Invisibility: Detecting Objects Invisible in Color Images}

\author{%
  Yongxin Wang \\
  George Mason University\\
  \texttt{ywang51@gmu.edu} \\
  \And
  Duminda Wijesekera \\
  George Mason University\\
  \texttt{dwijesek@gmu.edu}
}

\begin{document}

\maketitle

\begin{abstract}

Despite recent success of object detectors using deep neural networks, their deployment on safety-critical applications such as self-driving cars remains questionable. This is partly due to the absence of reliable estimation for detectors' failure under operational conditions such as night, fog, dusk, dawn and glare. Such unquantifiable failures could lead to safety violations. In order to solve this problem, we created an algorithm that predicts a pixel-level invisibility map for color images that does not require manual labeling - that computes the probability that a pixel/region contains objects that are invisible in color domain, during various lighting conditions such as day, night and fog. We propose a novel use of cross modal knowledge distillation from color to infra-red domain using weakly-aligned image pairs from the day and construct indicators for the pixel-level invisibility based on the distances of their intermediate-level features. Quantitative experiments show the great performance of our pixel-level invisibility mask and also the effectiveness of distilled mid-level features on object detection in infra-red imagery.

\end{abstract}

\section{Introduction}
\label{sec:intro}
Object detection has rapidly improved since the emergence of large scale data sets \cite{deng2009imagenet, everingham2010pascal, geiger2013vision, lin2014microsoft, yu2018bdd100k} and powerful baseline systems like two-stage detectors Fast/Faster/Mask R-CNN~\cite{girshick2015fast, ren2015faster, he2017mask} and one-stage detectors, such as YOLO~\cite{redmon2016you, redmon2017yolo9000, redmon2018yolov3}, SSD~\cite{liu2016ssd}, RetinaNet~\cite{lin2017focal}. 
One issue of importance in deploying detectors in safety-critical applications such as self-driving cars is the required high confidence ensuring that navigable regions are free of obstructing objects during operational weather and lighting conditions.
Failing to detect objects(false negatives) or provide warning signals, for example when crossing pedestrians or parking vehicles are left unnoticed by object detectors carries potentially disastrous consequences. 
While the performance of object detectors is improving, they cannot be guaranteed never to make mistakes\cite{8968525}. Thus reliable vision systems should account for "knowing when they don't know" besides just delivering high detection accuracy.
Our work addresses this problem by predicting a so called \emph{pixel-level invisibility map for color images} without manual labeling. 
Equipped with such invisibility map, a system could decide to trust detection results of some regions over others in an image or signal warning messages as well as hand over to a human. 

\begin{figure}
  \centering
  \includegraphics[width=\textwidth]{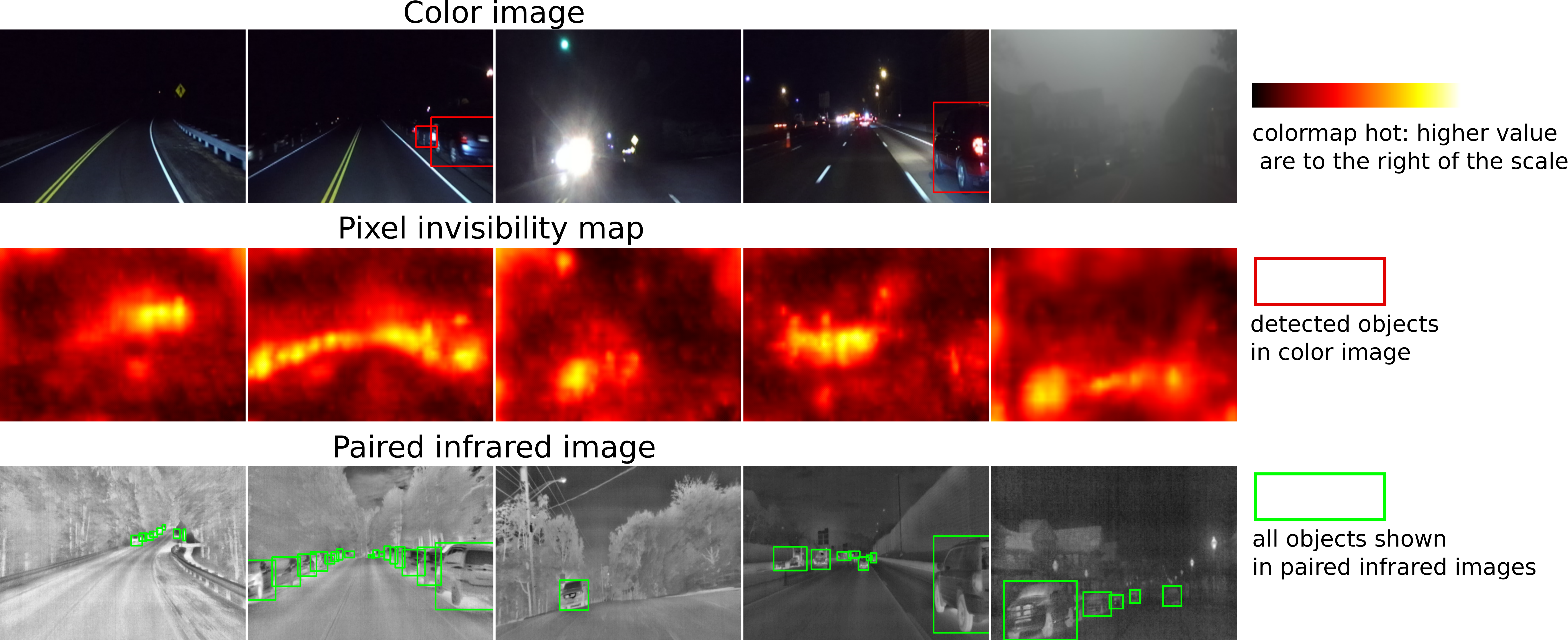}
  \caption{Our method predicts an invisibility map for each pixel (2nd row) given only the color image (1st row). Left to right: It can handle distant objects, highly occluded and cluttered objects, objects with strong glare, multiple objects from complex scenes like working zones on highway and objects in the fog. Color detectors can only detect objects (bounding boxes in 1st row) that are visible to its spectrum, while there are many more objects in the scene (3rd row). This could cause disastrous consequences where safety is crucial if such missed detection is trusted. Our pixel-level invisibility map indicates how much the detection results from color images may be trusted,  both for false negatives and false positives.}
  \label{fig:intro_problem}
\end{figure}

We define an \emph{invisibility mask} for one image as the likelihood that a region or pixel contains objects invisible in that domain. That is, the likelihood of one pixel or region contributing to false negatives in object detectors. Regions of color images during good daylight obtain low invisibility scores because visible light of enough energy is reflected to the camera by objects on the spot. Though dark regions of images in the night or obscure regions of images in the fog will have high invisibility scores. One straightforward approach to do this is to create a large labeled data set where every pixel in the image is labeled with an invisibility score, which is often expensive and ambiguous to collect. Instead, our method predicts the invisibility score for every pixel in the image without laborious labeling efforts by proposing a novel use of cross modal knowledge distillation and the generation of well-aligned image pairs between color and infra-red images.

Cross modal knowledge distillation~\cite{gupta2016cross} is also called supervision transfer as a means of avoiding labeling large scale data sets for certain modalities. Given paired images from two modalities, intermediate-level representations are transferred from richly annotated color domain to other modalities with limited labeled data sets such as depth images~\cite{gupta2016cross}, sound~\cite{aytar2016soundnet} and infra-red images. The novelty of our invisibility mask is in utilizing the supervision transfer from color to infra-red images of the daytime and then using distances between mid-level representations of two modalities to approximate perceptual distances between invisibility of two modalities in various lighting conditions including dusk, dawn, dark nights and fog. 

Knowledge distillation~\cite{gupta2016cross} specifically requires that the two modalities are presented in a paired fashion, especially in well-aligned manner for object detection and also for our pixel-level invisibility mask. Here, well-aligned image pairs are the ones where the corresponding pixels in the paired images are located at the same positions in their respective image planes. The raw image pairs captured by color and infra-red sensors have large displacements~\cite{zhang2019weakly, li2018multispectral}, which come from (1) Internal camera attribute differences such as focal length, resolution and lens distortion, (2) External transformations like pose differences and (3) Time differences from exposure time and shutter speed. We address the first two disparities by image registration using a known pattern board, while we propose \textbf{Align}ment \textbf{G}enerative \textbf{A}dversarial \textbf{N}etwork (AlignGAN) to alleviate the remaining displacements.

The contributions of our work is in three folds. (1) To the best of our knowledge, this is the first work to generate pixel-level invisibility mask for color images without manual labeling, hence contributing towards the failure detection problem caused by sensory false negatives in autonomous driving; (2) The direct transfer of mid-level representations from color image to infrared image gets promising detection accuracy in infra-red domain; (3) Mitigating misalignment problems(AlignGAN) present in color-infrared pairs. Extensive experiments are conducted to quantitatively evaluate the performance of our system.


\section{Related Work}
\label{section:related_work}
Our pixel-level invisibility mask is closely connected the task of uncertainty estimation~\cite{gal2016dropout, gal2016uncertainty, blundell2015weight, kendall2015bayesian, kendall2017uncertainties}, failure detection~\cite{daftry2016introspective, zhang2014predicting, hecker2018failure, ramanagopal2018failing, kuhn2020introspective, corbiere2019addressing} and out-of-distribution detection~\cite{devries2018learning, hendrycks2016baseline, liang2017enhancing, lee2017training}. Most reported works either consider at raw sensor data as introspective inspection~\cite{daftry2016introspective}, estimate the uncertainty of model-based classifiers or compute a confidence score or a binary decision. Our system differs from them by (1) Estimate the confidence of the sensor itself from the an outsider's viewpoint, which is the infra-red camera; (2) Predicting a confidence probability for every pixels in the image. 

Pixel objectness~\cite{jain2017pixel} is the first work to compute pixel-level masks for all \emph{object-like} regions. ~\cite{8968525} proposes a failure detection system for traffic signs where excited regions are extracted from feature maps in object detectors to narrow down both the manual labeling space and searching space for false negatives. Though they still need to label the excited regions as false negatives or true negatives. Our work also predicts pixel-level masks for all regions of potentially invisible objects(false negatives) in color images. In contrast, our method doesn't require any manual labeling of the invisible regions, and instead color-infrared image pairs are utilized to provide such supervision.

In order to get aligned color infra-red image pair, ~\cite{hwang2015multispectral} created \emph{KAIST Multispectral Pedestrian Dataset} based on a beam splitter to split a beam of light in two for color and infra-red cameras. Though image pairs in the data set were observed to have distinct displacements~\cite{zhang2019weakly, li2018multispectral} and also more affected in the night because of the intensity decay from the beam splitter. ~\cite{choi2018kaist} reports less severe disparity problems, though their data sets are not released yet.  
We observe the same problem in our data set which we collected using the setup of vertically aligned cameras and propose AlignGAN to mitigate such displacements in image planes.


Pix2pix~\cite{isola2017image} and CycleGan~\cite{zhu2017unpaired} developed methods for cross domain translation in paired and unpaired settings. Cycle consistency~\cite{zhu2017unpaired} is the main technique  to address unpaired cross-domain translation. Though these translations mainly address style transfers from source to target domains without moving the pixels in the source domain. Our AlignGAN module uses edge map learned from the infrared image to generate aligned color images.

The remainder of this paper describes our method and experiments in detail. In Section~\ref{section:system}, we present our system and describe the network architectures along with the training procedure to generate pixel-level invisibility map. Finally in Section~\ref{section:experiment} we conclude with extensive experiments on our data set and show several comparison results. Code, data, and models will be released.
\section{System Overview}
\label{section:system}

\begin{figure}
  \centering
  \includegraphics[width=\textwidth]{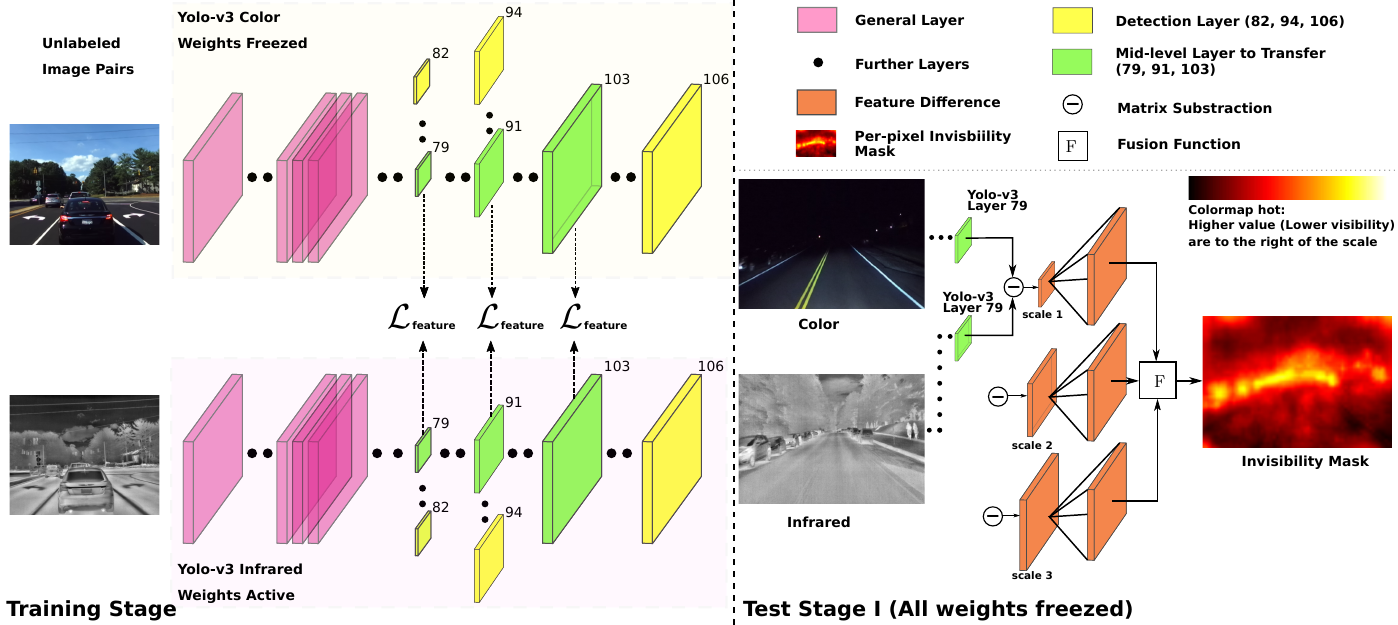}
  \caption{System overview. Our system takes well-aligned image pairs from unlabeled video stream. In the training stage(left side), the Yolo-v3 color network is freezed, while Yolov3-infrared network is trained to reproduce the mid-level features as the ones in Yolov-v3 color network at three scales, which are located at layer 79, 91 and 103 in the network. In the test stage(right), the difference of those features from layer 79, 91 and 103 are computed and fused to get the final invisibility mask.}
  \label{fig:system_overview}
\end{figure}

We present a system to learn to generate pixel-level invisibility maps for color images in an unsupervised way. During the training phase, our system takes weakly-aligned color-infrared image pairs that are looking at the same scene as input. Such imperfectly-aligned image pairs are firstly registered to remove the geometric differences and then aligned by AlignGAN to remove the remaining displacements between two modalities as detailed in Section~\ref{section:image_alignment}. After the image pairs are well-aligned, our Knowledge Transfer System transfers the learned representations from color domain to infrared domain. Then at test stage, the pairs of representations are compared to estimate the invisibility score of every pixel in the color image. As a side product, the learned representations of infrared images can be directly used to construct an object detector for infrared images without any manual labelling or retraining, as shown in Figure~\ref{fig:system_overview}.

\subsection{Alignment Generative Adversarial Network}
\label{section:image_alignment}

\begin{wrapfigure}{L}{0.6\textwidth}
\centering
\includegraphics[width=0.6\textwidth]{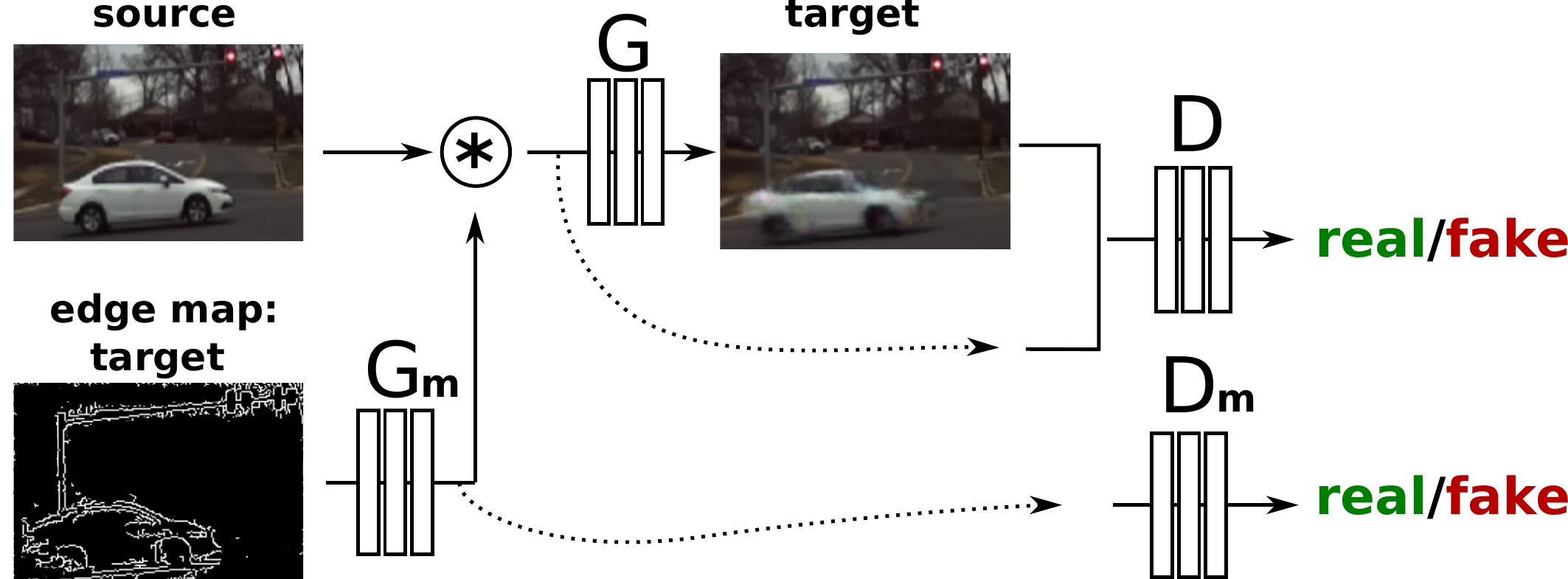}
\caption{\label{fig:alignment_module}{Base alignment network. The module takes as input source image and edge map from target position, and outputs the target image in the given position. The conversion from source to target which is performed using network $G$ is conditioned on the motion cues which are generated using network $G_m$. The target prediction after network $G$ and motion cues generated by $G_m$ are fed into two discriminators $D$ and $D_m$ respectively.}}
\end{wrapfigure}

\paragraph{AlignGAN:} Given the raw image pairs which are poorly aligned, we remove the internal and external transformations between two cameras using standard camera calibration technique, which we will describe in Section \ref{section:experiment}. Then we propose AlignGAN based on Generative Adversarial Network (GAN) to learn to generate the well-aligned color image from weakly-aligned color-infrared pairs. The base block of the systems is shown in Figure~\ref{fig:alignment_module}. As the figure shows, we use two streams to learning - both using the alignment block - during one iteration of the training phase. The first stream uses a color image  and a weakly-aligned infrared image as the edge map as the source, and produces a color image created using Flownet2~\cite{reda2017flownet2} as the target. In the other stream, source is image is still color image, edge map though is from a close color image $I_{c1}$ in the video stream, and the target image is the color image $I_{c1}$ itself.

We built our system based Pix2pix~\cite{isola2017image}. Both the generative network $G$ and $G_m$ use the U-Net architecture~\cite{ronneberger2015u} with an input size of $512 \times 512$.

\subsection{Knowledge Transfer From Color to Infrared Domains}
\label{section:knowledge_transfer}
We use YOLO-V3 architecture~\cite{redmon2018yolov3} to run the experiment mainly based three reasons. (1) It is fast and deployable in real-time applications like self-driving cars. (2) No need of a proposal network such as in the Faster RCNN and MASK-RCNN. (3) It has three detection modules based on three different image scales with good detection capabilities for small objects that may appear on a road.

We transfer the representations from the color domain to the infrared domain.  We choose the learned mid-level features of mages to transfer in the Yolo-V3 architecture that occurs prior to the detection stage. This is refereed to as  the \emph{mid-only} transfer. They are the outputs of layer 79, 91 and 103, respectively from three different scales, as shown in the left side of Figure~\ref{fig:system_overview}. Similar to~\cite{gupta2016cross}, we conducted two comparative experiments by transferring the last-layer detection results (Yolo-only) and both intermediate-level features and the detection results (Mid+Yolo). The detection results are computed based on the outputs of detection layers which are 82, 94 and 106.

We construct an infrared detector without any manual labeling and fine-tuning by concatenating learned intermediate features with the detection module from color detector, which produces promising detection accuracy for infrared images in different lighting conditions, as shown in the experiment Section~\ref{section:transfer_learning}.

\subsection{Invisibility Estimation}

We have now two detectors ${Yolo}_{c}$ and ${Yolo}_{i}$ where ${Yolo}_{c}$ is pre-trained on richly-annotated data sets and ${Yolo}_{i}$ is trained using well-aligned image pairs to reproduce the same intermediate-level features as the ones of their peer color images. Based on the observation that mid-level features for infra-red images are much less affected by the lighting conditions than the ones in color images, as shown in Section~\ref{section:transfer_learning}, we used the feature differences to estimate the lighting conditions and thereby estimated the visibility of every pixels in color images.

The YOLO-v3 architecture has three detection modules to estimate the at three different scales, and consequently it provides intermediate-level features at three different scales. We propose an invisibility score to integrate the features differences at different scales as shown on the right side of the Figure~\ref{fig:system_overview}. Here we define the invisibility score for an pixel $s_i$ as a function $F$ of the L2-distances of the mid-level features $\{d_k \mid k = 1, 2, 3\} $ between color and infra-red images in equation \ref{equation:score}. Here $d_1$, $d_2$, $d_3$ are from layer 79, 91, 103 respectively and $t_k$ is the highest value that we choose for $d_k$. Finally we trained a convolutional neural network based on U-Net~\cite{ronneberger2015u} to generate the invisibility mask even in the absence of infrared images.
\begin{equation}
\label{equation:score}
F(d_1, d_2, d_3) = 1 - \frac{1}{3} \sum\limits_{k=1}^3 min((t_k - d_k)/t_k, 0)
\end{equation}
\begin{figure}
  \centering
  \includegraphics[width=0.95\textwidth]{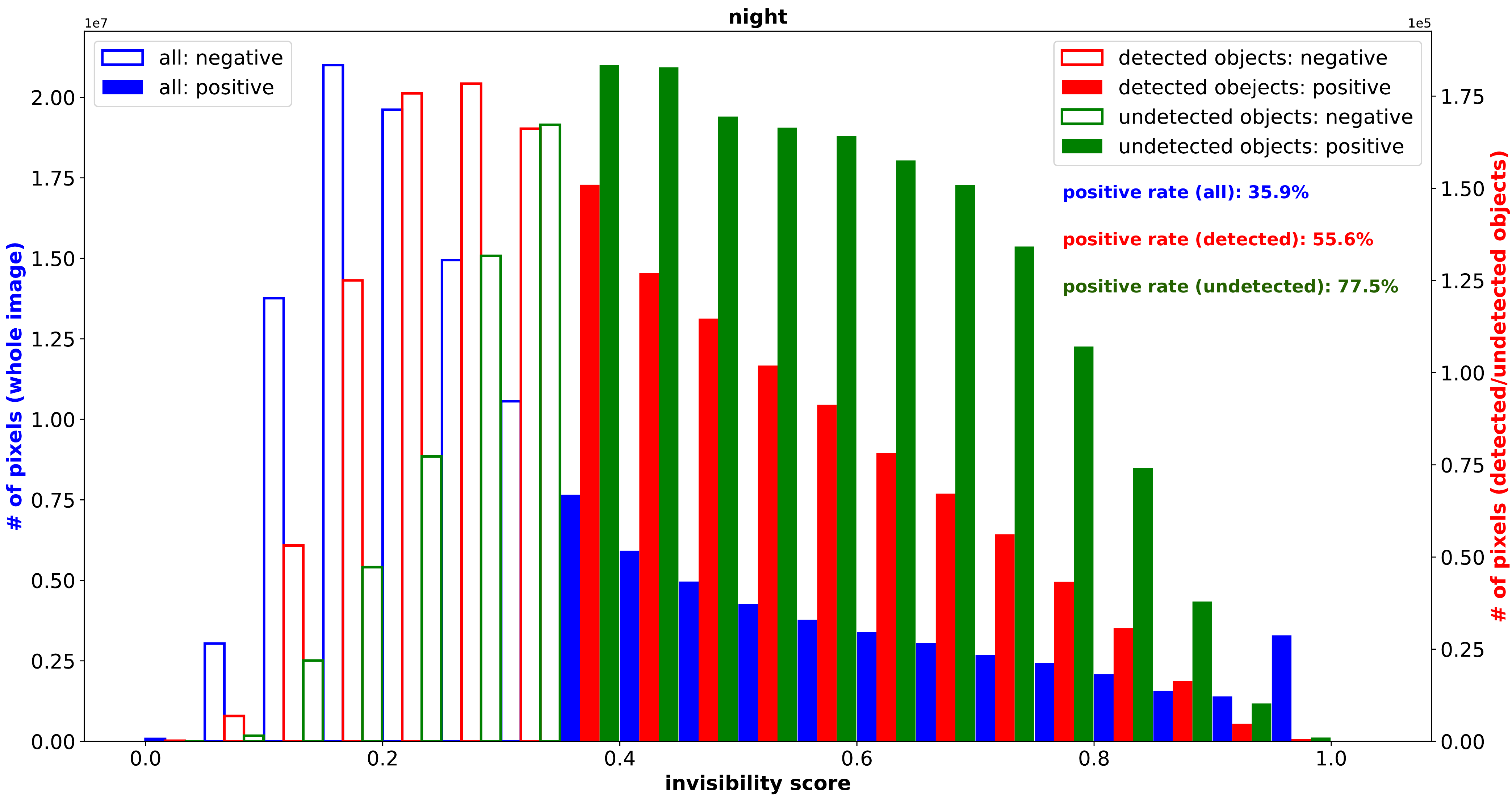}
  \caption{Histogram of Distances: We show the result of pixel-level invisibility mask on capturing the false negatives of MASK-CNN for night scenes. With a threshold of 0.35(separated by empty bars and solid bars) for invisibility score, the invisibility map can cover 77.5\% of the pixels in the undetected objects(green bar) while only take up 35.9\% of the pixels in the night images.}
  \label{fig:experiment_histogram}
\end{figure}

\section{Experiments}
\label{section:experiment}

\paragraph{Experimental Setup:} This section presents experimental outcomes for predicting undetectable areas in color images and the unsupervised knowledge transfer from color to infrared domain. We build a sensor platform that can be installed on the roof rack of any car (and we used such as setup for experiments). We used a FLIR ADK camera and the right ZED camera as our sensor pair. The color-IR calibration were performed using Caltech Calibration tools with a planar checkerboard partly made by aluminium foil. This removes major parts of camera distortion and help establish a coarse estimation for common field of view and scale differences. With Homographic warping based on pairs of corresponding points from two image planes, the disparity problem in static scene can be addressed well. Such weakly-aligned pairs of images are then taken as the input of AlignGAN. 

\paragraph{Data set - Color-IR Pair: } We sampled 18546 color-infrared pairs from the videos (around 120000 image pairs) that we collected while driving in the day to construct the training set for transferring the intermediate-level features from color domain to infra-red domain. For the validation data set, we collected and manually labeled 2000 image pairs with object bounding boxes, 500 during dawn, 500 during dusk, 500 during night and 500 during fog. They were used to evaluate the prediction performance of the undetectable area in color images and detection performance for infrared images. We don't have the exact statistics of our training set since we didn't label them by hand. Though our manually labeled validation set which contains 7821 cars, 1138 traffic signs, 626 persons and 343 trucks could give a clue to that of our training data set.

\paragraph{Experiment Focus:} We focus on answering the following five questions in this section. (1) How good is the prediction of invisible area in color images? (2) How good is the detection performance on infrared images through knowledge transfer? (3) Which level of representation transfer will give the best result on the detection accuracy? (4) Will our AlignGAN enhance the knowledge transfer process? and finally (5) How will the transfer performance change with respect to the number of images pairs? Now we answer those questions quantitatively using our results.

\subsection{How good is the prediction of the undetected area in the color image?}
\label{section:see_undetectable}

We use intermediate-level features from two paired DNNs as a space where Euclidean distance serves as the estimation of the reliability of color images compared to infra-red images. Our experiment result show that the proposed system can produce good quality masks for invisibility. In our experiment, we set the $t_1$, $t_2$, $t_3$ to 4, 3.5, 3.2 respectively in the Equation \ref{equation:score}.

\begin{wrapfigure}{L}{0.57\textwidth}

\centering

\includegraphics[width=0.57\textwidth]{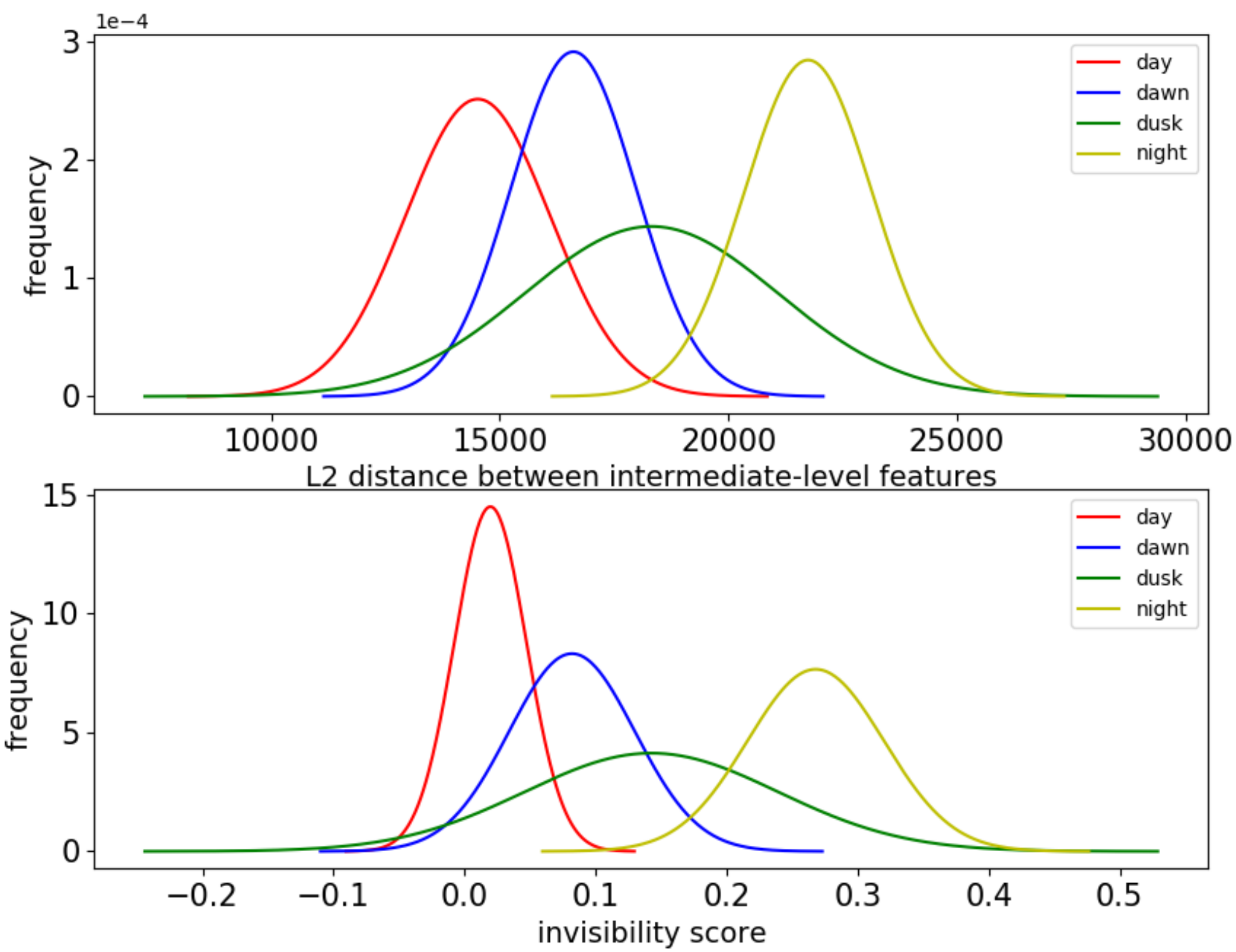}
\caption{\label{fig:exp_dist}{Distributions of distances and invisibility score. We show the L2 distance distributions (upper figure) between intermediate-level features from layer 79, 91 and 103 in Yolo-v3 and the invisibility scores (lower) from different lighting conditions. The Gaussian distributions from left to right are respectively from data of daytime, dawn, dusk and night}}
\end{wrapfigure}

For image pairs respectively from day, dawn, dusk, and night, we compute the L2 distance between their intermediate-level features and the invisibility score learned from our system. As shown in Figure~\ref{fig:exp_dist}, both the feature difference and invisibility score increase while the light intensity of the environment decreases. This is consistent with the observation that color images are more reliable in better lighting. Such agreement is validated by the different reactions to light change inherited in different spectra used in color and infrared cameras, which will be explained in detail in Section~\ref{section:transfer_learning}. The Gaussian distribution of invisibility score for day is (0.020, 0.028) and for night is (0.268, 0.052). These measures are highly separable with few overlaps in the distributions, as shown in Figure~\ref{fig:exp_dist}. These quantities show that invisibility score can be used to estimate the per-pixel invisibility in the color images with different lighting conditions.

A good binary visibility mask has two characteristics of (1) covering most of the undetectable objects (2) covering only the undetectable area in an image, which we use in quantified form to assess the effectiveness of our results. Now we report the visibility rate for all dawn, dusk and night scenes, where only the one for night scene is shown in Figure~\ref{fig:experiment_histogram},



Now we report results of a quantitative analysis showing the effectiveness of predicting the undetectable area in the color images. In Figure \ref{fig:experiment_histogram}, the distribution of the undetectable areas and the whole image are  different (mean 0.61 vs 0.27), indicating that the score is good representation of undetectability. The invisibility mask can cover 77.5\% of the undetectable area in the image with the visibility threshold of 0.35 and only report 35.9\% of the whole images. The invisibility mask for dawn time covers 49.2\% of the undetected area while reports 15.4\% of the whole image. And for dusk time, it covers 73.4\% of the undetected area while reports only 22.2\% of the image.

\subsection{Can paired data facilitate detection using transfer learning?}
\label{section:transfer_learning}

\begin{table}[t]
        \begin{minipage}{0.45\textwidth}
            \centering
             \caption{Detection accuracy for different lighting conditions. Even for night scenes which are not present in training set, the overall accuracy is up to 34.2\%}
            \begin{tabular}{lllll}
            \toprule
            Name            & Car       & Person    & \makecell{Traffic\\Light} & All   \\
            \midrule
            day         & \textbf{0.620}     & 0.191     & \textbf{0.576}         & \textbf{0.462} \\
            dusk        & 0.610     & 0.186     & 0.485         & 0.424 \\
            dawn        & 0.600     & \textbf{0.337}     & 0.476         & 0.471 \\
            night       & 0.506     & 0.120     & 0.399         & 0.342 \\
            fog         & 0.496     & 0.149     & 0.365         & 0.337 \\
            \bottomrule
          \end{tabular}
         
          \label{table:day night transfer}
        \end{minipage}
        \hfillx
        \begin{minipage}{0.5\textwidth}
            \centering
             \caption{Detection accuracy for different layers. We found that using only the mid-level features to transfer can achieve the best accuracy for object detection.}
            \begin{tabular}{lllll}
                \toprule
                Name            & Car       & Person    & \makecell{Traffic\\Light} & All   \\
                \midrule
                \textbf{M}id-only        & \textbf{0.513}     & \textbf{0.241}     & 0.470         & \textbf{0.408} \\
                \textbf{M}+\textbf{Y} 0.05   & 0.470     & 0.178     & \textbf{0.516}         & 0.388 \\
                \textbf{M}+\textbf{Y} 0.1    & 0.474     & 0.171     & 0.500         & 0.382 \\
                \textbf{M}+\textbf{Y} 1.0    & 0.437     & 0.170     & 0.475         & 0.361 \\
                \textbf{Y}olo-only       & 0.447     & 0.164     & 0.463         & 0.358 \\
                \bottomrule
              \end{tabular}
             
              \label{table:layers_comparison}
        \end{minipage}
    \end{table}


Firstly, we report that the knowledge transfer through mid-level features can reach 46.2\% overall detection accuracy for infra-red imagery in the Color-IR Pair data set. Since the learning doesn't require any manual annotations and doesn't require any retraining, we found the result to be promising. Again, this shows the importance of creating paired image for color and infrared images. Other than color-depth, color-pose and image-sound pairs, the color-infrared pairs resemble each other in appearances and textures. Our result here can be used for object detection, segmentation for infrared images, and can provide an alternative to laborious manual labeling  direct transfer of mid-level features from color to infrared imagery.

We evaluated the effectiveness of the AlignGAN using the post application of object detection. We tested the detection performance of the daytime data trained infrared detector on the night time data, and observe that it can still get to an overall accuracy of 34.2\%. The images used in training phase were chosen from day time with good lighting condition and the test set includes dawn, dusk, night and fog time imagery. Without any manually labelled training data, the detection IOU of cars can reach 50.6\% as shown in Table~\ref{table:day night transfer}. This quantity shows that the intermediate-level features of infrared images can be transferred smoothly from day to night, in contrast to the ones in color images. The principal used in our invisibility score is that \emph{features in IR is stable to light change and are trained to be like the ones in color, the mid-level features produced by infrared images are of same the caliber of features in color images when lighting conditions are poor}. This is the primary  reason for success on estimating the invisibility of the color images using our invisibility scores as shown in Section~\ref{section:see_undetectable}.

\subsection{Which layer is more effective?}

We experimented with the knowledge transfer from different layers. Gupta et al.~\cite{gupta2016cross} indicates that combining mid-level features with last layer features will give the best detection results when retraining on the target data sets. Although we show that using mid-level features only gives the best detection result of (40.8\%) over mid-last layer transfer (36.1\%) and the one using the yolo-layer only gives (35.8\%). These results are summarized in Table~\ref{table:layers_comparison}. 

More importantly, we varied the weight of the yolo layer to 0.05 and 0.1 in the loss function and conducted two more experiments. Surprisingly, we found that higher weights on the yolo layer resulted in worse overall detection ac curacies. One potential explanation for this observation is that the data set is not large enough to train a new object detector for all modules, especially for the class prediction and bounding box regression. Consequently, we learnt that it is more efficient to learn only the mid-level features and to not change layers of the detection module.

\begin{table}
  
  \centering
  \caption{Detection accuracy for AlignGAN. With the alignment module, there is relative 2.77\% boost in terms of overall accuracy.}
  \begin{tabular}{lllll}
    \toprule
    Name            & car       & Person    & traffic light & All   \\
    \midrule
    Mid+Yolo 1.0    & 0.437     & 0.170     & 0.475         & 0.361 \\
    Mid+Yolo 1.0 + Flow Encoder  & \textbf{0.442}     & \textbf{0.191}     & \textbf{0.481} & \textbf{0.371} \\
    \bottomrule
  \end{tabular}
  
  \label{table:flow_comparison}
\end{table}
\subsection{How much will the AlignGAN help the detection?}

\begin{figure}
\begin{minipage}[b]{0.5\textwidth}
\includegraphics[width=\textwidth]{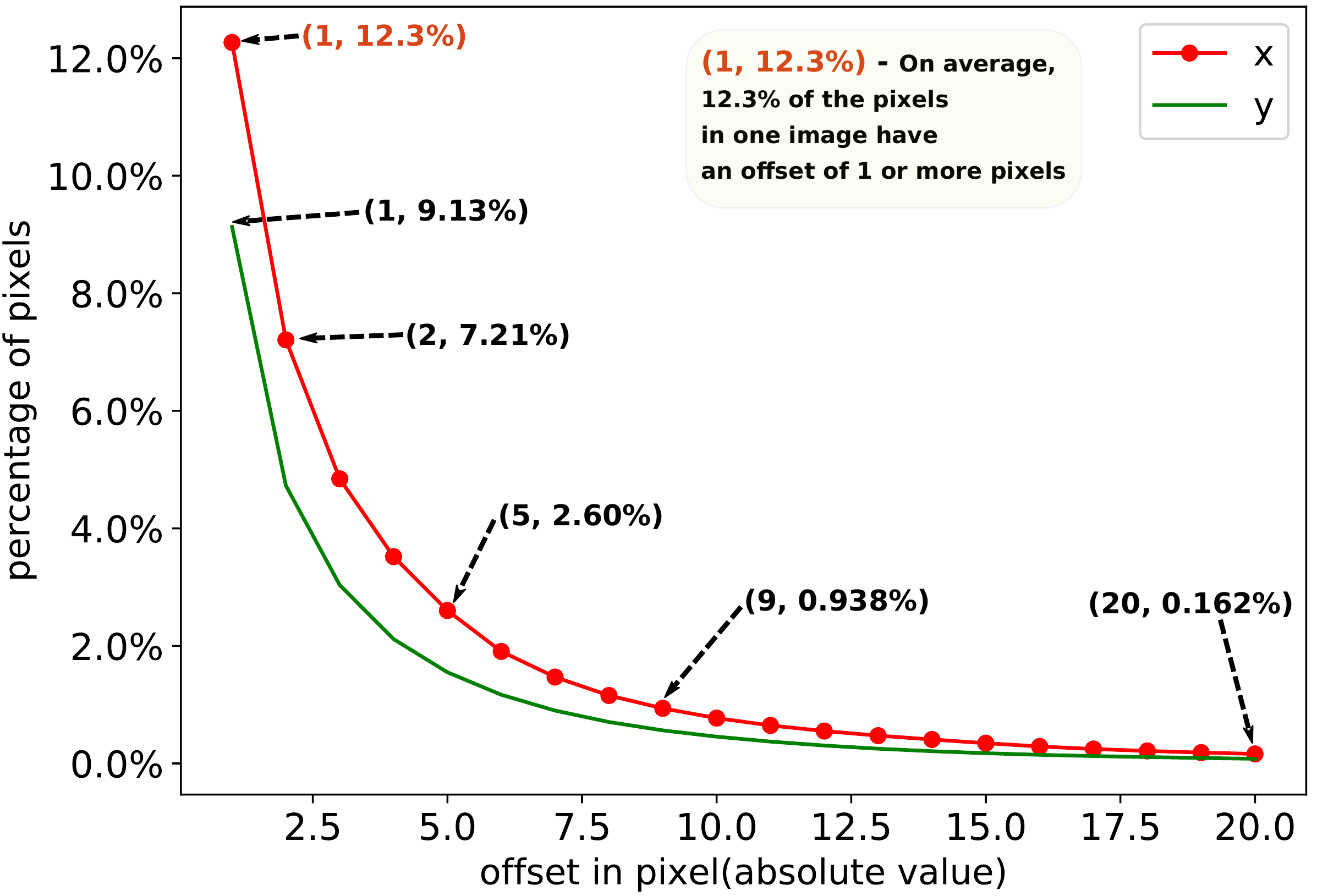}
\caption{\label{fig:exp_xy_offset}{X-Y offset distribution. We shows that in image plane offset in x direction is on average larger than that in y direction. Also distinct movements(5 pixels or more) constitute 2.6\% of the pixels in the images.}}
\end{minipage}
\hfill
\begin{minipage}[b]{0.47\textwidth}
\includegraphics[width=\textwidth]{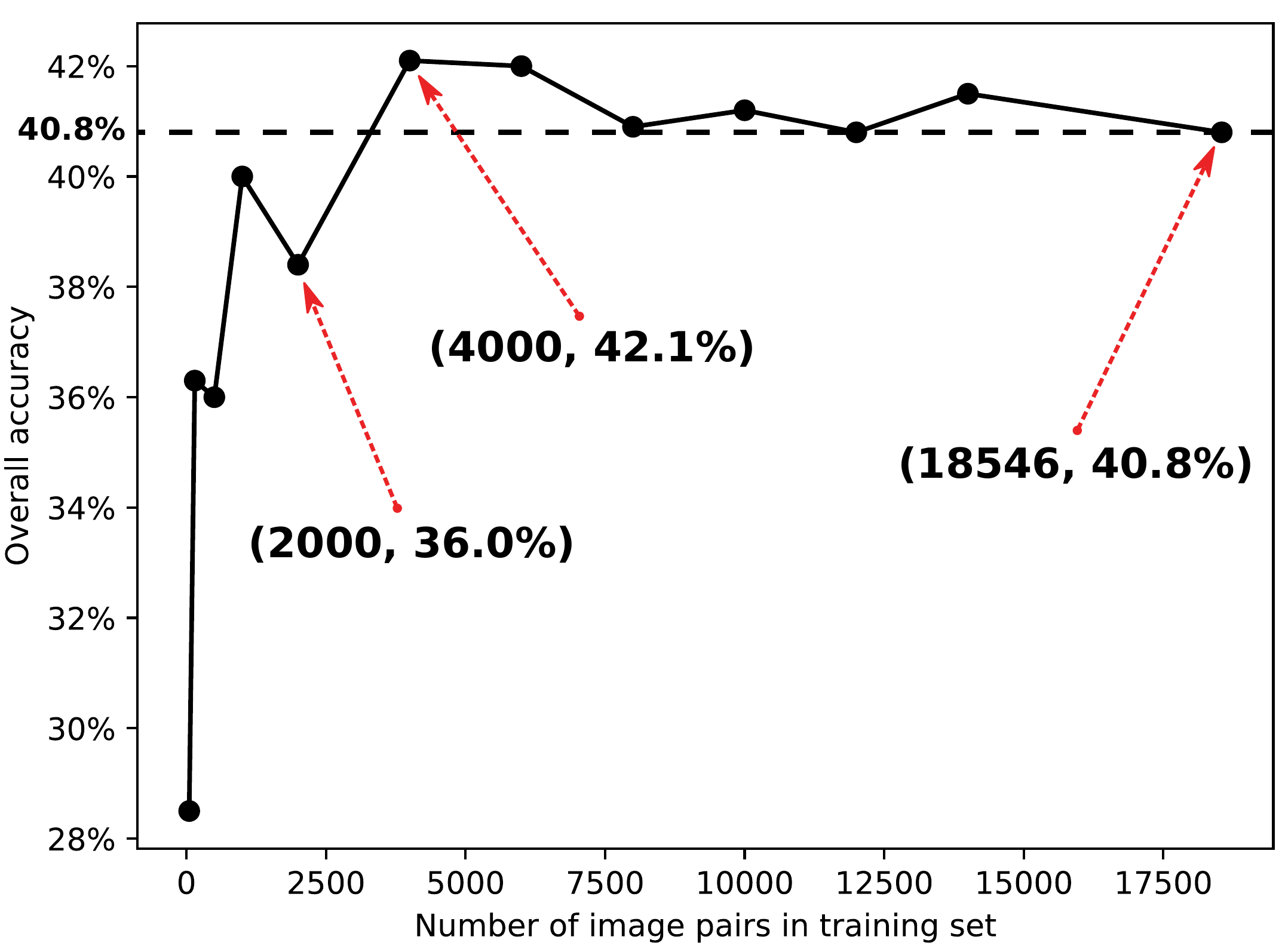}
\caption{\label{fig:size_dataset}{Comparison of data set sizes. We experiment with different sizes of training data sets for knowledge transfer, and observes that using 4000 image pairs can obtain the same if not better accuracy as the original size of 18,546 image pairs.}}
\end{minipage}
\end{figure}

Even after the pre-processing, the image pairs still have some displacements. Here we calculate the statistics of such displacements. With a image resolution of 640 by 512, the estimated X-Y displacement is shown in Figure~\ref{fig:exp_xy_offset}. On average, there are only about 2.6\% pixels in one image have 5 or more displacement on the X-direction, which we consider to be the threshold of movement that affect the detection results. Also, we noticed more displacements in the X-direction than in the Y-direction in the image plane. We attribute this to the fact object movements  projected to image plane is more obvious in x-direction than the y-direction. 

We evaluated how AlignGAN can improve the data transfer from color domain to infrared domain. Our Alignment module is based on the Pix2pix network~\cite{isola2017image} and FlowNet 2~\cite{ilg2017flownet} and are shown in Table~\ref{table:flow_comparison} showing an enhancement of 2.77\% within the AlignGAN module.

\subsection{How many pairs are needed to get a good transfer?}
Section~\ref{section:knowledge_transfer} showed that mid-level feature transfer results in the best performance. We now determine the number of image pairs required to achieve that performance. Surprisingly,  Figure \ref{fig:size_dataset} show that randomly sampled 4000 image pairs from the space of 18546 image pairs can achieve the same if not better accuracy than the entire sample space. This observation implies that the domain difference from color to infrared can be learned from a small amount images pairs and the transfer discriminating visual representations from the well-established color detection task to infrared images can be done in a light-weighted manner.
Figure ~\ref{fig:size_dataset} shows experiments with different number images from 100 to 18000 and that the performance will be stable after 4000 images. Such result may appear to be counter intuitive at first sight as more data often leads to better results when modal capacity is large enough like the one we use in the experiment. One potential explanation for the saturation point is that the two domains have much in common and thus the domain difference can be mitigated with a few examples. This saturation point observation with 4000 samples can be used as a promising deployment strategy of direct knowledge transfer. Consequently, less time is needed to trained for both invisibility system and the direct knowledge transfer for object detection.

\section{Conclusion}
\label{sec:conclusions}

Given a color image, we predict a pixel-level invisibility mask for such image without manual labelling. Experiments have shown that our mask is able to distinguish invisible pixels from the visible pixels. Our results also demonstrate the effectiveness of building an object detector for the infrared domain using the mid-level features transferred from its peer color images. Our pixel-level invisibility mask can also be used as the confidence map to fuse the results from multiple sensors.

\begin{ack}
\end{ack}

\section*{Broader Impact}
In safety critical applications like self-driving cars, false negatives of object detection is the bottleneck of its deployment on perceptual tasks. Leaving objects of concern such as pedestrians and vehicles undetected could lead to potentially disastrous damages. Our system is able to predict the invisible objects in color images in the form of pixel-level invisibility map and estimate the reliability of object detectors with respect to the sensor's limitation. Equipped with such invisibility map, a system could decide to trust the decisions made on sensor input or signal warning message along with hand over to a human. This is a huge step forward to ensure safety for perceptual systems based on color cameras. Also the visibility mask will be useful in sensor fusion as the confidence map, which will improve the overall accuracy for object detection and tracking in autonomous vehicles. At certain cases, our system may also miss a false negative from object detectors. As no single failure prediction method can detect all failure cases, we strongly encourage the fusion of various failure detection algorithms on safety-critical applications. Finally, we didn't observe biases in the data.


\bibliographystyle{splncs04}
\bibliography{neurips_2020}

\begin{thebibliography}{10}
\providecommand{\url}[1]{\texttt{#1}}
\providecommand{\urlprefix}{URL }
\providecommand{\doi}[1]{https://doi.org/#1}

\bibitem{aytar2016soundnet}
Aytar, Y., Vondrick, C., Torralba, A.: Soundnet: Learning sound representations
  from unlabeled video. In: Advances in neural information processing systems.
  pp. 892--900 (2016)

\bibitem{blundell2015weight}
Blundell, C., Cornebise, J., Kavukcuoglu, K., Wierstra, D.: Weight uncertainty
  in neural networks. arXiv preprint arXiv:1505.05424  (2015)

\bibitem{choi2018kaist}
Choi, Y., Kim, N., Hwang, S., Park, K., Yoon, J.S., An, K., Kweon, I.S.: Kaist
  multi-spectral day/night data set for autonomous and assisted driving. IEEE
  Transactions on Intelligent Transportation Systems  \textbf{19}(3),  934--948
  (2018)

\bibitem{corbiere2019addressing}
Corbi{\`e}re, C., Thome, N., Bar-Hen, A., Cord, M., P{\'e}rez, P.: Addressing
  failure prediction by learning model confidence. In: Advances in Neural
  Information Processing Systems. pp. 2898--2909 (2019)

\bibitem{daftry2016introspective}
Daftry, S., Zeng, S., Bagnell, J.A., Hebert, M.: Introspective perception:
  Learning to predict failures in vision systems. In: 2016 IEEE/RSJ
  International Conference on Intelligent Robots and Systems (IROS). pp.
  1743--1750. IEEE (2016)

\bibitem{deng2009imagenet}
Deng, J., Dong, W., Socher, R., Li, L.J., Li, K., Fei-Fei, L.: Imagenet: A
  large-scale hierarchical image database. In: 2009 IEEE conference on computer
  vision and pattern recognition. pp. 248--255. Ieee (2009)

\bibitem{devries2018learning}
DeVries, T., Taylor, G.W.: Learning confidence for out-of-distribution
  detection in neural networks. arXiv preprint arXiv:1802.04865  (2018)

\bibitem{everingham2010pascal}
Everingham, M., Van~Gool, L., Williams, C.K., Winn, J., Zisserman, A.: The
  pascal visual object classes (voc) challenge. International journal of
  computer vision  \textbf{88}(2),  303--338 (2010)

\bibitem{gal2016uncertainty}
Gal, Y.: Uncertainty in deep learning. University of Cambridge  \textbf{1}, ~3
  (2016)

\bibitem{gal2016dropout}
Gal, Y., Ghahramani, Z.: Dropout as a bayesian approximation: Representing
  model uncertainty in deep learning. In: international conference on machine
  learning. pp. 1050--1059 (2016)

\bibitem{geiger2013vision}
Geiger, A., Lenz, P., Stiller, C., Urtasun, R.: Vision meets robotics: The
  kitti dataset. The International Journal of Robotics Research
  \textbf{32}(11),  1231--1237 (2013)

\bibitem{girshick2015fast}
Girshick, R.: Fast r-cnn. In: Proceedings of the IEEE international conference
  on computer vision. pp. 1440--1448 (2015)

\bibitem{gupta2016cross}
Gupta, S., Hoffman, J., Malik, J.: Cross modal distillation for supervision
  transfer. In: Proceedings of the IEEE conference on computer vision and
  pattern recognition. pp. 2827--2836 (2016)

\bibitem{he2017mask}
He, K., Gkioxari, G., Doll{\'a}r, P., Girshick, R.: Mask r-cnn. In: Proceedings
  of the IEEE international conference on computer vision. pp. 2961--2969
  (2017)

\bibitem{hecker2018failure}
Hecker, S., Dai, D., Van~Gool, L.: Failure prediction for autonomous driving.
  In: 2018 IEEE Intelligent Vehicles Symposium (IV). pp. 1792--1799. IEEE
  (2018)

\bibitem{hendrycks2016baseline}
Hendrycks, D., Gimpel, K.: A baseline for detecting misclassified and
  out-of-distribution examples in neural networks. arXiv preprint
  arXiv:1610.02136  (2016)

\bibitem{hwang2015multispectral}
Hwang, S., Park, J., Kim, N., Choi, Y., So~Kweon, I.: Multispectral pedestrian
  detection: Benchmark dataset and baseline. In: Proceedings of the IEEE
  conference on computer vision and pattern recognition. pp. 1037--1045 (2015)

\bibitem{ilg2017flownet}
Ilg, E., Mayer, N., Saikia, T., Keuper, M., Dosovitskiy, A., Brox, T.: Flownet
  2.0: Evolution of optical flow estimation with deep networks. In: Proceedings
  of the IEEE conference on computer vision and pattern recognition. pp.
  2462--2470 (2017)

\bibitem{isola2017image}
Isola, P., Zhu, J.Y., Zhou, T., Efros, A.A.: Image-to-image translation with
  conditional adversarial networks. In: Proceedings of the IEEE conference on
  computer vision and pattern recognition. pp. 1125--1134 (2017)

\bibitem{jain2017pixel}
Jain, S.D., Xiong, B., Grauman, K.: Pixel objectness. arXiv preprint
  arXiv:1701.05349  (2017)

\bibitem{kendall2015bayesian}
Kendall, A., Badrinarayanan, V., Cipolla, R.: Bayesian segnet: Model
  uncertainty in deep convolutional encoder-decoder architectures for scene
  understanding. arXiv preprint arXiv:1511.02680  (2015)

\bibitem{kendall2017uncertainties}
Kendall, A., Gal, Y.: What uncertainties do we need in bayesian deep learning
  for computer vision? In: Advances in neural information processing systems.
  pp. 5574--5584 (2017)

\bibitem{kuhn2020introspective}
Kuhn, C., Hofbauer, M., Petrovic, G., Steinbach, E.: Introspective black box
  failure prediction for autonomous driving. In: 31st IEEE Intelligent Vehicles
  Symposium (2020)

\bibitem{lee2017training}
Lee, K., Lee, H., Lee, K., Shin, J.: Training confidence-calibrated classifiers
  for detecting out-of-distribution samples. arXiv preprint arXiv:1711.09325
  (2017)

\bibitem{li2018multispectral}
Li, C., Song, D., Tong, R., Tang, M.: Multispectral pedestrian detection via
  simultaneous detection and segmentation. arXiv preprint arXiv:1808.04818
  (2018)

\bibitem{liang2017enhancing}
Liang, S., Li, Y., Srikant, R.: Enhancing the reliability of
  out-of-distribution image detection in neural networks. arXiv preprint
  arXiv:1706.02690  (2017)

\bibitem{lin2017focal}
Lin, T.Y., Goyal, P., Girshick, R., He, K., Doll{\'a}r, P.: Focal loss for
  dense object detection. In: Proceedings of the IEEE international conference
  on computer vision. pp. 2980--2988 (2017)

\bibitem{lin2014microsoft}
Lin, T.Y., Maire, M., Belongie, S., Hays, J., Perona, P., Ramanan, D.,
  Doll{\'a}r, P., Zitnick, C.L.: Microsoft coco: Common objects in context. In:
  European conference on computer vision. pp. 740--755. Springer (2014)

\bibitem{liu2016ssd}
Liu, W., Anguelov, D., Erhan, D., Szegedy, C., Reed, S., Fu, C.Y., Berg, A.C.:
  Ssd: Single shot multibox detector. In: European conference on computer
  vision. pp. 21--37. Springer (2016)

\bibitem{8968525}
{Rahman}, Q.M., {Sünderhauf}, N., {Dayoub}, F.: Did you miss the sign? a false
  negative alarm system for traffic sign detectors. In: 2019 IEEE/RSJ
  International Conference on Intelligent Robots and Systems (IROS). pp.
  3748--3753 (2019)

\bibitem{ramanagopal2018failing}
Ramanagopal, M.S., Anderson, C., Vasudevan, R., Johnson-Roberson, M.: Failing
  to learn: autonomously identifying perception failures for self-driving cars.
  IEEE Robotics and Automation Letters  \textbf{3}(4),  3860--3867 (2018)

\bibitem{reda2017flownet2}
Reda, F., Pottorff, R., Barker, J., Catanzaro, B.: flownet2-pytorch: Pytorch
  implementation of flownet 2.0: Evolution of optical flow estimation with deep
  networks (2017)

\bibitem{redmon2016you}
Redmon, J., Divvala, S., Girshick, R., Farhadi, A.: You only look once:
  Unified, real-time object detection. In: Proceedings of the IEEE conference
  on computer vision and pattern recognition. pp. 779--788 (2016)

\bibitem{redmon2017yolo9000}
Redmon, J., Farhadi, A.: Yolo9000: better, faster, stronger. In: Proceedings of
  the IEEE conference on computer vision and pattern recognition. pp.
  7263--7271 (2017)

\bibitem{redmon2018yolov3}
Redmon, J., Farhadi, A.: Yolov3: An incremental improvement. arXiv preprint
  arXiv:1804.02767  (2018)

\bibitem{ren2015faster}
Ren, S., He, K., Girshick, R., Sun, J.: Faster r-cnn: Towards real-time object
  detection with region proposal networks. In: Advances in neural information
  processing systems. pp. 91--99 (2015)

\bibitem{ronneberger2015u}
Ronneberger, O., Fischer, P., Brox, T.: U-net: Convolutional networks for
  biomedical image segmentation. In: International Conference on Medical image
  computing and computer-assisted intervention. pp. 234--241. Springer (2015)

\bibitem{yu2018bdd100k}
Yu, F., Xian, W., Chen, Y., Liu, F., Liao, M., Madhavan, V., Darrell, T.:
  Bdd100k: A diverse driving video database with scalable annotation tooling.
  arXiv preprint arXiv:1805.04687  (2018)

\bibitem{zhang2019weakly}
Zhang, L., Zhu, X., Chen, X., Yang, X., Lei, Z., Liu, Z.: Weakly aligned
  cross-modal learning for multispectral pedestrian detection. In: Proceedings
  of the IEEE International Conference on Computer Vision. pp. 5127--5137
  (2019)

\bibitem{zhang2014predicting}
Zhang, P., Wang, J., Farhadi, A., Hebert, M., Parikh, D.: Predicting failures
  of vision systems. In: Proceedings of the IEEE Conference on Computer Vision
  and Pattern Recognition. pp. 3566--3573 (2014)

\bibitem{zhu2017unpaired}
Zhu, J.Y., Park, T., Isola, P., Efros, A.A.: Unpaired image-to-image
  translation using cycle-consistent adversarial networks. In: Proceedings of
  the IEEE international conference on computer vision. pp. 2223--2232 (2017)

\end{thebibliography}
\end{document}